\pdfoutput=1

\documentclass[11pt]{article}

\usepackage[]{acl}

\usepackage{times}
\usepackage{arydshln}

\usepackage{wrapfig}
\usepackage{url}
\usepackage{latexsym}
\usepackage{color}
\usepackage{booktabs}
\usepackage{subfigure} 
\usepackage{multirow}
\usepackage{verbatim}
\usepackage{amsfonts}
\usepackage{float}
\usepackage{bm}
\usepackage{comment}
\usepackage{array}
\usepackage{graphicx}  
\usepackage{amsmath}
\usepackage{blindtext}
\usepackage[T1]{fontenc}

\usepackage[utf8]{inputenc}

\usepackage{microtype}

\title{HopPG: Self-Iterative Program Generation for Multi-Hop Question Answering over Heterogeneous Knowledge }

\author{Yingyao Wang\textsuperscript{1}\thanks{~~The first two authors contribute equally to this work.}, Yongwei Zhou\textsuperscript{1}\footnotemark[1],  Chaoqun Duan\textsuperscript{2}, Junwei Bao\textsuperscript{2}, Tiejun Zhao\textsuperscript{1} \\
    \textsuperscript{1}Harbin Institute of Technology \;\;  \textsuperscript{2}JD AI Research \\
    \{yywang, ywzhou\}@hit-mtlab.net \;\;  baojunwei001@gmail.com  \;\; tjzhao@hit.edu.cn
    }

\begin{document}
\maketitle
\begin{abstract}
The semantic parsing-based method is an important research branch for knowledge-based question answering.
It usually generates executable programs lean upon the question and then conduct them to reason answers over a knowledge base.
Benefit from this inherent mechanism, it has advantages in the performance and the interpretability.
However, traditional semantic parsing methods usually generate a complete program before executing it, which struggles with multi-hop question answering over heterogeneous knowledge. On one hand, generating a complete multi-hop program relies on multiple heterogeneous supporting facts, and it is difficult for generators to understand these facts simultaneously. On the other hand, this way ignores the semantic information of the intermediate answers at each hop, which is beneficial for subsequent generation. 
To alleviate these challenges, we propose a self-iterative framework for multi-\textbf{hop} \textbf{p}rogram \textbf{g}eneration (HopPG) over heterogeneous knowledge, which leverages the previous execution results to retrieve supporting facts and generate subsequent programs hop by hop. We evaluate our model on MMQA-T$^2$\footnote{A subset of MMQA, we detailed introduce the MMQA-T$^2$ dataset in Section~\ref{intro} and Section~\ref{dataset}.}, and the experimental results show that HopPG outperforms existing semantic-parsing-based baselines, especially on the multi-hop questions.
\end{abstract}
\section{Introduction\label{intro}}
Question answering is a fundamental task and plays a crucial role in the field of natural language processing \cite{cui2020mutual,kwiatkowski2019natural,liu2020logiqa,choi2018quac,fan2019eli5}. 
In recent years, question-answering tasks based on heterogeneous knowledge (HQA) have increasingly gained the attention of researchers \cite{2020HybridQA,2020Open,2021TAT,2021MultiModalQA,2021FinQA}. These tasks require models to perform multi-hop reasoning on different structured knowledge, i.e., tables and texts. One category of the existing HQA method performs implicit answer reasoning, which takes the question and knowledge as input and performs reasoning in the semantic space, and then directly outputs the answer \cite{2020Unsupervised,2021Iterative,2022MuGER,2021MATE,2021Multi}. Although this approach has proven effective in HQA tasks, it lacks interpretability, scalability, and symbolic reasoning abilities. In contrast, the semantic parsing-based (SP-based)  approach explicitly derives answers by generating and executable programs, remedying the above-mentioned deficiencies and enabling researchers to monitor and improve each step of the derivation process.

SP-based methods have been widely used in question-answering tasks on homogeneous knowledge sources, such as tables, knowledge graphs, and texts \cite{yih2014semantic,bao2014knowledge,bao2016constraint,abdelaziz2021semantic,zhou2022opera}. Nevertheless, the SP-based question-answering methods over heterogeneous knowledge still require further exploration. Recently, \citet{zhou-etal-2022-unirpg} introduced UniRPG, a SP-based model designed for HQA. 
They defined a set of general atomic and higher-order operations for discrete reasoning over heterogeneous knowledge resources. During the program generation process, UniRPG takes questions and supporting facts as input pairs of BART \cite{lewis-etal-2020-bart} to generate a complete program in a single step.

Although UniRPG has the ability to generate programs with tables and passages as supporting facts, it still struggles with multi-hop question answering for the following two reasons.
First, generating a complete multi-hop program usually depends on multiple heterogeneous supporting facts, making it challenging for the model to receive and understand all the facts simultaneously due to the length limitation.
On the other hand, intuitively, the reasoning results from the current hop are useful for selecting supporting facts and program generation in the next step. 
However, generating a complete program sequence before executing ignores the interaction between the reasoning results of the current hop and the subsequent program generation.
To tackle these issues, we introduce HopPG, an iterative program generation framework designed explicitly for multi-hop answer reasoning based on heterogeneous knowledge. HopPG leverages the execution results of the previous program to select supporting facts and generate subsequent programs iteratively. In comparison to UniRPG, HopPG reduces the knowledge complexity used in each program generation step and incorporates information from the previous steps, enhancing the model's capability for multi-hop reasoning. 

In this paper, we utilize a subset of the MMQA dataset \cite{2021MultiModalQA} for evaluating HopPG. Specifically, we only focus on questions based on tables and texts, which we refer to as MMQA-T$^2$, excluding those requiring images as knowledge. 
It possesses the following notable characteristics: 1) questions are based on heterogeneous knowledge, 2) questions require multi-hop reasoning, and 3) MMQA-T$^2$ provides detailed annotations of the supporting facts and intermediate results for each hop, allowing us to construct more accurate pseudo programs. The experimental results show that HopPG outperforms existing SP-based baselines, especially on the multi-hop questions.

Our contributions are summarized as follows: 
\begin{itemize}
    \item We propose HopPG, a self-iterative program generation framework  for multi-hop question answering over heterogeneous knowledge. 
    The framework successfully addresses the limitations of existing SP-based models. 
    \item We collect the MMQA-T$^2$ dataset based on MMQA, which only contains multi-hop questions over tabular and textual knowledge. Moreover, we construct pseudo-multi-hop programs for each question in MMQA-T$^2$ to train the program generator with weak supervision.
    \item We conduct extensive experiments and ablation studies on the MMQA-T$^2$ dataset. The experimental results show that HopPG outperforms the existing SP-based QA model, especially on the multi-hop questions.
\end{itemize}
\section{Related Work}
\subsection{Question Answering over Heterogeneous Knowledge}
Previous works have attempted to leverage knowledge with various background knowledge for question-answering, e.g., knowledge graphs and texts \cite{sun2019pullnet,han-etal-2020-open}. 
However, these researches are limited to using one type of knowledge as auxiliary information during answer reasoning and have not fully explored multi-hop reasoning across heterogeneous knowledge (HQA). To fill this gap, \citet{2020HybridQA} first propose the HybridQA dataset, which provides a WiKiTable  accompanied by hyperlinked Wikipedia passages as evidence for each question.
Based on HybridQA, \citet{chen2021ottqa} proposed the OTT-QA dataset, requiring the system to retrieve relevant tables and texts for their given question. Additionally, \citet{2021TAT} and \citet{2021FinQA} introduced TAT-QA and FinQA, both requiring numerical reasoning over heterogeneous data.

\subsection{Semantic Parsing-based Methods}
The semantic parsing-based methods reason answers by translating questions into executable programs such as SPARQL \cite{xiong2022autoqgs} and SQL\cite{hui-etal-2022-s2sql}. Previous semantic-parsing-based question-answering methods always research over homogeneous knowledge resources, i.e., texts \cite{zhou2022opera}, tables \cite{liu2022tapex} and knowledge graphs \cite{yih-etal-2016-value}. \citet{zhou-etal-2022-unirpg} proposed UniRPG, which first applies the semantic parsing-based method on question answering over heterogeneous knowledge including tables and texts. Inspired by these works, we propose HopPG in this paper to address the limitations of UniRPG in generating multi-hop programs based on heterogeneous knowledge.
\begin{figure*}[t]
	\centering
	\includegraphics[width=\textwidth]{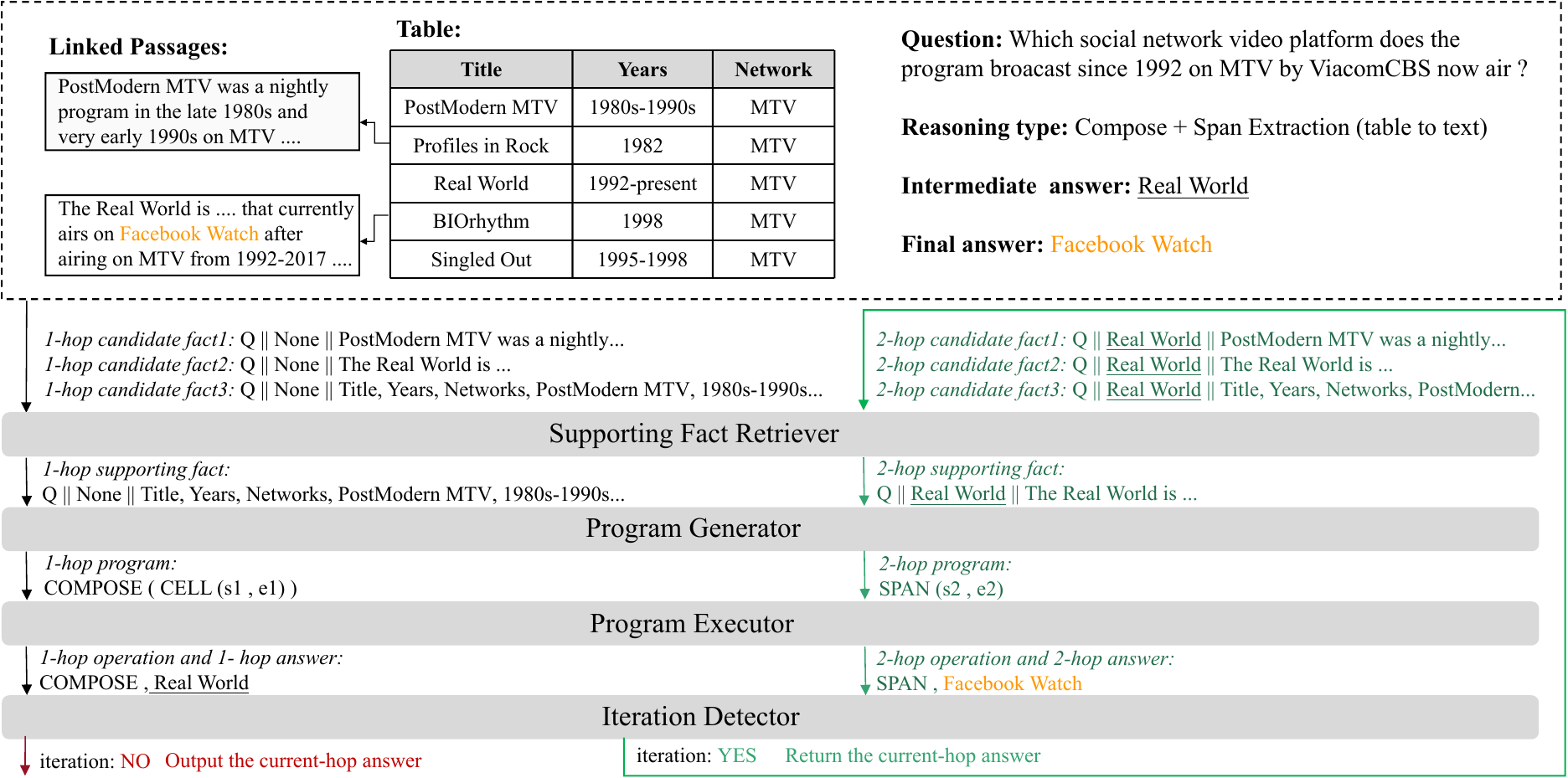}
	\caption{The framework of  HopPG. This figure gives a 2-hop question as the example.}
	\label{model}
\end{figure*}

\section{Methodology}
\subsection{Task Definition}
Multi-hop question answering over heterogeneous knowledge, i.e., tables and texts, aims to retrieve the supporting facts from the given knowledge and derive the answer to the question. 
Given a $H$-hop question $\mathcal{Q}$ and $K$ fact candidates $\mathcal{T}=\{t^1,...t^K\}$, the model requires deriving the final answer $a^H$ from $\mathcal{T}$. Apart from $a^H$, we use $\{a^1,...,a^{H-1}\}$ to represent the intermediate reasoning results in each hop. Each $a^h$, where $h \leq H$, can be a table cell, a text span, or a math calculation result.  The task is formalized as follows: 
\begin{align}
    a^H = \arg \max P(a^H|\mathcal{Q},\mathcal{T};\theta).
\end{align}
 In this work, $\mathcal{T}$ contains a table and a set of texts. We assume that $t^1$ represents the table and the others $\{t^2,...,t^K\}$ are the $K$-$1$ texts. Specifically, the table $t^1$ consists of $m \times n$ cells $\{c_{ij}\}$, where $m$ and $n$ are the numbers of rows and columns. 

 \begin{table*}[t]
\small
\centering
\begin{tabular}{l|l|l}
\toprule
\multirow{2}{*}{\textbf{Reasoning Type} } &
\multicolumn{2}{c}{\textbf{Multi-hop Program Templates}} \\\cline{2-3}
 &HOP-1 & HOP-2\\

\midrule
\multirow{5}{*}{Span Extraction} 
& $\mathtt{CELL}$ $(s,e)$& -\\
& $\mathtt{CELL\_VALUE}$ $(s,e)$& -\\
& $\mathtt{SPAN}$ $(s,e)$& -\\
& $\mathtt{SPAN\_VALUE}$ $(s,e)$& - \\
& $\mathtt{MULTISPAN( CELL_1 / SPAN_1, ..., CELL_m / SPAN_m )}$& - \\
\midrule
YesNo &$\mathtt{YESNO}$ $(s,e)$& -\\
\midrule
\multirow{2}{*}{Compare} 
& $\mathtt{ARGMAX (KV_1, KV_2)}$& -\\
& $\mathtt{ARGMIN (KV_1, KV_2)}$& -\\
\midrule
\multirow{4}{*}{Calculation}
& $\mathtt{SUM ( CV_{1}, ..., CV_{m} )}$ & -\\
&$\mathtt{AVG ( CV_{1}, ..., CV_{m} )}$& -\\
& $\mathtt{COUNT ( CV_{1}, ..., CV_{m} )}$& -\\

\midrule
Compose + Span Extraction & $\mathtt{COMPOSE(CELL / SPAN)}$ & $\mathtt{CELL / SPAN / MULTISPAN }$  \\
Compose + YesNo & $\mathtt{COMPOSE(CELL / SPAN)}$ & $\mathtt{YESNO}$\\
\midrule

Intersect	&$\mathtt{INTERSECT (MULTISPAN)}$ & $\mathtt{MULTISPAN}$\\

\bottomrule
\end{tabular}
\caption{The defined multi-hop program templates for MMQA-T$^2$.
}
\label{temp}
 
\end{table*}
\subsection{Definition of Multi-hop Program\label{op}}
\subsubsection{Operations}
HopPG generates logical programs based on the pre-difined operations. Aligning with the previous work \cite{zhou-etal-2022-unirpg}, our operations contains 4 atomic operations and 11 high-order operations, including ${\mathtt{CELL}}$,
${\mathtt{SPAN}}$,
${\mathtt{CELL\_VALUE}}$, ${\mathtt{SPAN\_VALUE}}$, ${\mathtt{KV}}$, ${\mathtt{MULTISPAN}}$, ${\mathtt{COUNT}}$, ${\mathtt{SUM}}$, ${\mathtt{AVG}}$, ${\mathtt{ARGMAX}}$, and ${\mathtt{ARGMIN}}$. Based on these operations, we extend an atomic operation and two high-order operations to enable the programs to efficiently solve multi-hop questions and more reasoning types. The extended operations are listed as follows:
\begin{itemize}
    \item ${\mathtt{YESNO}}$: This is an atomic operation designed for "yes or no" reasoning. HopPG converts this reasoning type as a test span extraction problem. Specifically, we connect two words, "yes" and "no", at the beginning of the input sequence of the generator. The arguments of ${\mathtt{YESNO}}$ are the same as the other atomic operations, which are $(s, e)$. During execution, ${\mathtt{YESNO}}$ checks the extracted span, if its content is "no", ${\mathtt{YESNO}}$ returns "no" as the answer. For all other cases, the operation returns "yes". 
    
    \item ${\mathtt{COMPOSE}}$: This operation means the current hop is an intermediate hop, and the answering reasoning requires iteration. The argument of the ${\mathtt{COMPOSE}}$ is one of the atomic operations, ${\mathtt{CELL}}$ or ${\mathtt{SPAN}}$, and ${\mathtt{COMPOSE}}$ directly returns the result of the atomic operation as the current-hop result.
    
    \item ${\mathtt{INTERSECT}}$: This operation also means the current hop is an intermediate hop. Its argument is a high-order operation ${\mathtt{MULTISPAN}}$, and ${\mathtt{INTERSECT}}$ directly returns the results of   ${\mathtt{MULTISPAN}}$, a set of table cells or text spans.
\end{itemize}



\subsubsection{Program Templates}
We define multi-hop program templates for questions of different reasoning types. These templates are used to construct pseudo programs for weakly supervised training and to constrain program generation.  We decompose the multi-hop reasoning into multiple single-hop programs. In this work, we set the maximum hop count
$H$=2. The defined templates are listed in Table~\ref{temp}. Specifically, for questions of "compose" reasoning types, HopPG directly outputs the execution results of the 2-hop program as the final answer. In contrast, when tackling "intersect" questions, HopPG compares the results from the two hops and outputs the overlap cells or spans as the final answers. We execute the pseudo programs  constructed using our templates and evaluate the question-answering performance on the development set of MMQA-T$^2$, the EM and F1 scores are 91.27\% and 93.89\%, respectively.

\subsection{Framework of HopPG }
We first decompose H-hop reasoning into multiple single hops, and define the corresponding programs as $\mathcal{P}$=$\{p^1,...,p^H\}$. Based on our program templates, HopPG generates the program $p^h$ hop by hop in an iterative way.

As Figure~\ref{model} shows, the framework of our HopPG mainly contains three modules: \textit{fact retriever}, \textit{program generator}, \textit{program executor}, and a \textit{iteration detection process}. 
During $h$-hop reasoning, 
the \textbf{fact retriever} first selects a supporting fact $t^h$ for the current hop based on the question $Q$ and the previous-hop result $a^{h-1}$. After that, the \textbf{program generator} receives $Q$, $a^{h-1}$ and $t^h$ as inputs and generate the program $p^h$. Subsequently, $p^h$ is executed by the \textbf{program executor} and $a^h$ can be derived. 
Notably, for programs comprising multi-level operations, the executor executes from the atomic operations to the high-order operations.
At this point, the \textbf{iteration detection} process checks if the high-order operation of $p^h$ is a multi-hop operation\footnote{${\mathtt{COMPOSE}}$ and ${\mathtt{INTERSECT}}$, which are defined in Section~\ref{op}}. If it is, HopPG returns $a^h$ to iterate the above process. Otherwise, HopPG terminates the iteration and outputs $a^h$ as the final answer.

In HopPG, the fact retriever and the program generator are trainable and trained separately among these modules. We will introduce their details in the following sections.

\subsubsection{Supporting Fact Retriever}
The retriever in HopPG aims to select the supporting fact $t^h$ for the h-hop program generation  from the provided candidates, including a table and a set of texts. for the $h$-hop program generation. Following  \cite{yoran-etal-2022-turning}, we finetune the BART-large \cite{lewis-etal-2020-bart} model using the training set of MMQA-T$^2$ as our retriever. The input of the retriever is a sequence consisting of the question $Q$, the golden execution result of the previous hop $\Bar{a}^{h-1}$, and one of the fact candidates $t^i$:
\begin{equation}
    \text{Inp}^{h}_{R}=[\langle s \rangle;Q;\langle \backslash s \rangle;\Bar{a}^{h-1};\langle \backslash s \rangle;t^i]
\end{equation}
Notably, the table $t^1$ is flattened by connecting its rows. For a first-hop example, the previous-hop execution result $]\Bar{a}^0$ is set to "$\mathtt{None}$". The retriever receives the input sequences of all candidates and outputs a score vector $\bm \delta$=$(s^1,...,s^K)$ of them, then the model is fine-tuned using the following loss function, where $l$ is the golden fact's index:
\begin{equation}
{\mathcal{L}} = \mathtt{CrossEntropy}(l,\mathtt{Softmax}(\bm{ \delta}))
\end{equation}
After tuning, the supporting fact retrieval accuracy of our retriever is 90.7\%. During the program generation process, the tuned retriever ranks all candidate inputs and selects the fact $t^h$=$argmax(\bm \delta)$ as the supporting fact of the current hop $h$. 

\begin{table*}[t]
\centering
\small 
\begin{tabular}{c|c|c|c}
\toprule
\textbf{Knowledge} & \textbf{Question Type} & \textbf{Hop}& \textbf{Distribution} \\
\midrule
\multirow{6}{*}{Only Table}
& Span Extraction &1& 42.5\%\\
& YesNo &1& 3.1\% \\

& Compare &1& 5.8\% \\
& Calculation &1& 0.6\% \\
& Intersect &2& 2.8\%\\
& Compose + Span Extraction &2& 6.0\%\\
\midrule
\multirow{2}{*}{Only Text}
& Span Extraction &1& 10.7\%\\
& YesNo &1& 5.1\%\\
\midrule
\multirow{4}{*}{Table + Text}  &Intersect &2& 4.1\%\\ &Compose + Span Extraction (table to text) &2& 9.4\%\\
 &Compose + Span Extraction (text to table) &2& 4.8\%\\
 &Compose + YesNo (table to text) &2& 2.0\%\\

  &Compare (Compose + Span Extraction (text to table)) &3& 3.0\%\\
\bottomrule
\end{tabular}
\caption{The question type distribution of the training set of MMQA-T$^2$.
}
\label{questions} 
\end{table*}
\begin{table*}[t]
\centering
\small
\begin{tabular}{l|ccccccc|c}
\toprule
 & \bf Extraction & \bf Compare & \bf Compose & \bf YesNo & \bf Calculation & \bf Intersect & \bf Compose\_Compare & \bf Total\\
\midrule
Train &7512&819&3463&1170&101&978&425&14122\\
Dev &834&90&346&130&11&108&47&1566\\
Test &748&70&383&142&6&88&64&1501\\
\bottomrule
\end{tabular}
\caption{Basic statistics of MMQA-T$^2$.}\label{tab:dataset_sts}

\end{table*}

\subsubsection{Program Generator}
The program generator aims to generate the corresponding program for each hop. In this work,  the generator is a BART-based model equipped with a structure-aware knowledge reader \cite{zhang-etal-2020-table}, which is designed to enhance the table understanding ability of the encoder. We use the training set of MMQA-T$^2$ together with our constructed pseudo programs to train the UniRPG. 
Specifically, for  the $h$th-hop program generation, the input sequence consists of the question $Q$, the execution result of the previous-hop program $a^{h-1}$, and the supporting fact $t^h$ selected 
 by the retriever. We connect a text span \textit{"Yes or No"} to the question to transfer the $\mathtt{YESNO}$ reasoning type into a span extraction problem.
 Formally, the $h$th-hop input of our generator is represented as follows:
 \begin{equation}
    \text{Inp}^{h}_{G}=[\langle s \rangle;\textit{yes or no};Q;\langle \backslash s \rangle;a^{h-1};\langle \backslash s \rangle;t^h]
\end{equation}

The input sequence is fed into the structure-aware knowledge reader, which injects table structure information into the self-attention layers of the BART-encoder with structure-aware attention.
The reader learns the input and outputs the representations of input tokens.
Then, we feed it into the encoder to learn the representation $\mathbf{K} = \{\mathbf{k}_i\}_{i=1}^{L}$, where
$L$ is the length of the input sequence.

Subsequently, the representations vectors $\mathbf{K}$ of the input tokens are used to decode the program based on our defined operations. For the generator training, we collect the golden supporting fact and the corresponding program for each hop of the questions. All these data are utilized to train the program generator.

\subsubsection{Program Executor}
To perform answer derivation, we implement a symbolic program executor for HopPG that executes the generated programs based on their meanings. When dealing with programs comprising multiple levels of operations, the executor executes from the atomic operations to the high-order operations.

\subsubsection{Iteration Detector}
In HopPG, we add an iteration detection process after the program execution to determine whether further hop generation and reasoning are needed. During the inference phase of HopPG, the iteration detection process checks the outermost operation of the current generated program to determine if it is one of the multi-hop operations, which include $\mathtt{COMPOSE}$ and $\mathtt{INTERSECT}$. If the operation belongs to multi-hop operations, HopPG performs the next hop generation iteratively. Otherwise, the current result is considered as the final answer.
\begin{table*}[t]
\centering
\small

\begin{tabular}{l|ccc|ccc}
\toprule
\textbf{Model} & \multicolumn{3}{c|}{\textbf{MMQA}} & \multicolumn{3}{c}{\textbf{MMQA-T$^2$}}\\
\midrule
&Overall& Single-hop&Multi-hop &Overall& Single-hop& Multi-hop\\
\midrule
AutoRouting &  42.10 / 49.05 & - & -&-&-&-\\
Implicit-Decomp  & 48.80 / 55.49& 51.61 / 58.36 & 44.59 /51.19 & 54.30 / 62.15 & 57.34 / 64.75 & 46.23 / 55.26\\

UniRPG$^{\dagger}$ & 53.87 / 60.20 & 57.22 / 64.05  & 48.87 / 54.45 & 63.09 / 70.29 & 65.05 / 72.72 & 56.69 / 63.84 \\
\midrule
HopPG (ours) & \textbf{54.61} / \textbf{61.00} & \textbf{58.38} / \textbf{64.92} & \textbf{48.98} / \textbf{55.18} &\textbf{ 63.76} / \textbf{71.14} & \textbf{66.42} / \textbf{73.54} & \textbf{57.91} / \textbf{64.77}  \\
\bottomrule
\end{tabular}
\caption{Results on the complete MMQA and the MMQA-T$^2$ datasets.}
\label{tab:mainres}

\end{table*}
\begin{table*}[t]
\centering
\small 
\begin{tabular}{c|c|c|c}
\toprule
\textbf{Type} & \textbf{Implicit-Decomp} & \textbf{UniRPG$^{\dagger}$}& \textbf{HopPG} \\
\midrule

TableQ &\textbf{72.35} / \textbf{79.89} & 63.41 / 69.93 & 65.04 / 70.33\\
TextQ &49.65 / 50.01& 65.60 / 73.64 & \textbf{67.13} / \textbf{75.19}  \\
Compose(TableQ,TextQ)  &50.00 / 56.70& 70.73 / 75.80& \textbf{71.95} / \textbf{75.99} \\
Compose(TextQ,TableQ)  &37.96 / 49.89 & 50.93 / 60.67 & \textbf{56.48} / \textbf{68.05}\\
Intersect(TableQ,TextQ) &48.98 / 54.24 & 44.89 / 48.63 & \textbf{48.98} / \textbf{55.24} \\

\bottomrule
\end{tabular}
\caption{Results on questions with different types of MMQA-T$^2$. }
\label{tab:abres}
 
\end{table*}

\section{Experiments}

\subsection{Dataset\label{dataset}}
The MMQA-T$^2$  used in this paper is a subset of MMQA \cite{2021MultiModalQA}. Specifically, we collect all questions based on tables and texts and exclude questions using images from MMQA, in a total of 15688 training instances and 1501 development instances.  Each question in MMQA is provided with 1 table together with 10 texts as candidate facts. The question type of the collect instances is shown in Table~\ref{questions}. Based on the question types, we re-split the training instances we collected in a ratio of 9:1 as the training set of the development set of  MMQA-T$^2$, respectively. The collected development instances from MMQA are directly used as the test set of  MMQA-T$^2$. We give the final basic statistics of MMQA-T$^2$ in Table~\ref{tab:dataset_sts}. 

To further demonstrate MMQA-T$^2$, we present the distribution of the training set questions in Table~\ref{questions} based on knowledge utilized, reasoning type, and the number of hops. It can be observed that 38.5\% of the questions are multi-hop, with 23.3\% of those requiring both tables and text to derive answers. These statistics indicate that MMQA-T$^2$ is a suitable dataset for evaluating HopPG, designed to improve question-answering performance in multi-hop reasoning over heterogeneous knowledge.

\subsection{Implementation Details}
The program generator of HopPG is initialized using BART-base and optimized with AdamW. The training settings are consistent with UniRPG. Specifically, the learning rate, batch size and weight decay are set to 1e-4, 128 and 0.01, respectively. When generating programs, we set the Beam Size of the beam search method as 4. The experiments are conducted on NVIDIA A100 GPU.
\subsection{Baselines}
\paragraph{AutoRouting and Implicit-Decomp}  These two baselines are from \cite{2021MultiModalQA}. We compare HopPG's question-answering performance with these baselines on the original MMQA dataset. For questions requiring images, we directly use the prediction results from Implicit-Decomp.
\paragraph{UniRPG} We reproduce the UniRPG-base version to directly generate complete multi-hop programs based on our operations. We select the top 2 candidate knowledge ranked by retrieval scores as the supporting facts. Then the question and the two facts are concatenated as input for UniRPG. In addition, we cut off the over-length input directly.
\subsection{Main Results}
Table~\ref{tab:mainres} shows the question-answering results of our HopPG and the baseline models. We use EM and F1 scores as the evaluation metrics and report the results on both the original MMQA and MMQA-T$^2$ datasets. For image-based questions in MMQA, we use the predictions of Implicit-Decomp. As the table shows, UniRPG, the first semantic parsing-based method on HQA tasks, achieves significant performance on MMQA-T$^2$ by 63.09 EM and 70.29 F1. It is proved that UniRPG can effectively solve the answer reasoning over tabular and textual knowledge by generating executable programs. 

Based on the results of UniRPG, our HopPG further brings improvements on MMQA-T$^2$. The EM scores increase by 1.37 on the single-hop questions and 1.22 on the multi-hop questions. The improvements are from the ability of HopPG to reduce the complexity of the supporting facts, and sufficiently utilize previous-hop execution results during program generations. These improvements demonstrate the effectiveness of HopPG. 
\begin{table*}[ht]
\centering
\footnotesize
\resizebox{\textwidth}{!}{
\begin{tabular}{c|c|c|c}
\toprule
& \textbf{Model} & \textbf{Golden Program} &  \textbf{Generated Program}
\\
\midrule
\multirow{3}{*}{Q1} 
&UniRPG &$\mathtt{SPAN(86,89)}$&$\mathtt{SPAN(86,164)}$ \\
\cdashline{2-4}[1pt/1pt]
&HopPG-h1 & $\mathtt{SPAN(88,91)}$ & $\mathtt{SPAN(88,91)}$\\
&HopPG-h2 & - & -\\
\midrule

\multirow{3}{*}{Q2} 
&UniRPG &$\mathtt{CELL(244,248)}$&$\mathtt{MULTISPAN(CELL(58,69),...,CELL(244,264))
}$ \\
\cdashline{2-4}[1pt/1pt]
&HopPG-h1 & $\mathtt{COMPOSE(MULTISPAN(CELL(71,76),CELL(100,101)))
}$ & $\mathtt{COMPOSE(MULTISPAN(CELL(71,76),CELL(71,76))
)}$\\
&HopPG-h2 & $\mathtt{CELL(252,256)}$ & $\mathtt{CELL(252,256)
}$\\
\midrule

\multirow{3}{*}{Q3} 
&UniRPG &$\mathtt{CELL(352,356)}$&$\mathtt{CELL(352,373)
}$ \\
\cdashline{2-4}[1pt/1pt]
&HopPG-h1 & $\mathtt{COMPOSE(CELL(354,358))}$ & $\mathtt{COMPOSE(CELL(354,358))}$\\
&HopPG-h2 & $\mathtt{CELL(356,360)}$ & $\mathtt{CELL(356,360)}$\\

\bottomrule
\end{tabular}  
}
\caption{Case studies.
}
\label{tab:case}
 
\end{table*}
\subsection{Ablation Studies}
To provide a more detailed and intuitive demonstration of HopPG's performance on different question types, we compare the question-answering results between HopPG and baselines in Table~\ref{tab:abres}, where the results are reported on questions with different hop numbers and knowledge sources.
\subsubsection{Questions with Homogeneous Knowledge}
The results in the table indicate that Implicit-Decomp performs excellently on table-based questions, because it utilizes the table pre-trained model, TAPAS \cite{herzig-etal-2020-tapas}, for table-based question answering. Compared to TAPAS, semantic parsing-based models like UniRPG and HopPG offer the advantage of interpretability and avoid the need for expensive pre-training on a large number of tables. The table also shows that HopPG outperforms the baseline models on text-based questions because it generates programs using only the selected text as input, rather than simply concatenating complex and heterogeneous knowledge. This greatly reduces the difficulty of model inference.
\subsubsection{Multi-hop Questions with Heterogeneous Knowledge}
As expected, HopPG achieves significant improvements in multi-hop question answering, which are mainly from the following three reasons: 1) Compared to directly generating complete programs, HopPG improves the accuracy of program generation at each hop by reducing the complexity of knowledge. 2) The program generation for each hop can refer to the execution results of the previous hop. 3) In HopPG, the errors in the program generated at the previous hop do not directly lead to incorrect results in the final output, which to some extent reduces error propagation.

Compared to Implicit-Decomp, our semantic-parsing-based pipeline brings significant improvement on multi-hop questions, even without using the table pre-trained model for TableQ. This further proves the effectiveness of our pipeline designed for multi-hop and hybrid questions without fine-designation for table question answering.

\subsection{Case Studies}
\subsubsection{Q1: Fix the single-hop program.}
In Table \ref{tab:case}, we present three cases of questions that UniRPG fails to answer but were fixed by HopPG. Among them, Q1 is a single-hop question with the answer contained in a given textual knowledge. For UniRPG, all candidate texts are concatenated with flattened tables as input the sequence. In this experiment, the pseudo-program for Q1 is the same in both UniRPG and HopPG. As shown in the first row of the table, UniRPG incorrectly predicts the end index of the answer string in the serialized knowledge, while HopPG successfully extracts the answer. This is because UniRPG directly concatenates all candidate supporting facts as input, which makes it difficult for the model to understand the knowledge and reason about the answer due to the redundant information. In contrast, HopPG reduces the complexity of the input knowledge by retrieving the necessary supporting fact for each hop, thereby improving the reasoning accuracy.
\subsubsection{Q2: Fix the second hop of the multi-hop program.}
UniRPG cannot handle Q2, a multi-hop question, due to its limited operation set for multi-hop reasoning. Therefore, in UniRPG, Q2 is treated as a single-hop question and directly annotated with the span extraction pseudo-program. As shown in Table \ref{tab:case}, the model is confused and incorrectly predicts the  $\mathtt{MULTISPAN}$ operation for Q2 instead of $\mathtt{SPAN}$. because the question format of multi-hop questions is different from that of single-hop questions. The incorrect operation further leads to the generation of a series of wrong string indices. For simplicity, we omits the intermediate  $\mathtt{CELL}$ operations in UniRPG's result for Q2 in Table \ref{tab:case}. In contrast, HopPG generates two-hop programs iteratively for this question. Although the generated first-hop program is not entirely correct, the lack of information about an intermediate result does not have a decisive impact on the generation of the program for the second iteration. HopPG ultimately successfully generates the correct second-hop program for Q2 and obtains the correct answer.
\subsubsection{Q3: Fix all hops of the multi-hop program.}
Q3 is also a multi-hop question. As shown in the table, UniRPG predicts the correct operation, but it incorrectly predict the string index for the answer. In contrast, HopPG successfully generates a two-hop program for the question not only obtaining the correct answer but also making the reasoning process interpretable, demonstrating the advantage of HopPG in handling multi-hop questions.
\subsection{Error Analysis}

To conduct error analysis over the test set of MMQA-T$^2$, we collected the programs generated by HopPG that are inconsistent with the pseudo programs we annotated. These wrong programs are primarily caused by two reasons: incorrect operation prediction and incorrect string index $(s,e)$ prediction. According to our statistical analysis, among the wrong cases of single-hop questions, 27\% of them have incorrect operation predictions, while 99\% have incorrect string index predictions. The proportions of these two reasons in the cases of two-hop questions ('COMPOSE' reasoning type) also align with the aforementioned trend. Specifically, for their first-hop program generation, the proportions of incorrect operation prediction and incorrect string index prediction are 32.4\% and 100\%, respectively. For the second-hop, the proportions of these two factors are 7\% and 98\%, respectively. This indicates that if the operation selection is incorrect, the model will struggle to accurately retrieve the require information from the knowledge.

Additionally, in the wrong cases of two-hop questions, the proportions of errors in the first-hop and second-hop program generation are 50.0\% and 69.9\%, respectively. Based on our observations, an error in the first-hop program does not necessarily lead to an error in the second-hop generation, as the execution results of the first-hop program only serve as input information for the second-hop. In fact, among the cases where the first-hop program generation are incorrect, 60.2\% of them generate the correct second-hop programs and obtain the correct answers to the questions. This proves that the iterative generation way of HopPG can to some extent mitigate the impact of reasoning errors in previous hops on the final result.

\section{Conclusion}
We propose HopPG, a self-iterative program generation approach for multi-hop question answering over heterogeneous knowledge. Unlike directly generating complete programs for multi-hop questions, HopPG iteratively generates programs for each hop based on the execution results from the previous-hop program. We evaluate our model using a subset of MMQA that only contains text-based and table-based questions and construct pseudo programs for each instance to train HopPG under weak supervision. The experimental results demonstrate that HopPG brings significant improvements for multi-hop question answering over heterogeneous knowledge, outperforming semantic parsing-based question-answering models that directly generate complete programs.
\bibliography{anthology,custom}
\bibliographystyle{acl_natbib}

\appendix



\end{document}